\documentclass[sigconf]{acmart}

\usepackage{booktabs} 
\usepackage{times}
\usepackage{latexsym}
\usepackage{url}
\usepackage[utf8]{inputenc}
\usepackage{graphicx}
\usepackage{natbib}
\graphicspath{ {images/} }

\setcopyright{rightsretained}

\renewcommand\footnotetextcopyrightpermission[1]{} 
\begin{document}
\title{Extracting Fairness Policies from Legal Documents}

\author{Rashmi Nagpal}
\affiliation{
  \institution{IIIT Delhi}
}
\email{rashmi14085@iiitd.ac.in}

\author{Chetna Wadhwa}
\affiliation{
  \institution{IIIT Delhi}
}
\email{chetna17009@iiitd.ac.in}

\author{Mallika Gupta}
\affiliation{
  \institution{IIIT Delhi}
}
\email{mallika17024@iiitd.ac.in}

\author{Samiulla Shaikh}
\affiliation{
  \institution{IBM Research}
}
\email{samiullas@in.ibm.com}

\author{Sameep Mehta}
\affiliation{
  \institution{IBM Research}
}
\email{sameepmehta@in.ibm.com}

\author{Vikram Goyal}
\affiliation{
  \institution{IIIT Delhi}
}
\email{vikram@iiitd.ac.in}


\begin{abstract}

Machine Learning community is recently exploring the implications of bias and fairness with respect to the AI applications. The definition of fairness for such applications varies based on their domain of application. The policies governing the use of such machine learning system in a given context are defined by the constitutional laws of nations and regulatory policies enforced by the organizations that are involved in the usage. Fairness related laws and policies are often spread across the large documents like constitution, agreements, and organizational regulations. These legal documents have long complex sentences in order to achieve rigorousness and robustness. Automatic extraction of fairness policies, or in general, any specific kind of policies from large legal corpus can be very useful for the study of bias and fairness in the context of AI applications.

We attempted to automatically extract fairness policies from publicly available law documents using two approaches based on semantic relatedness. The experiments reveal how classical Wordnet-based similarity and vector-based similarity differ in addressing this task. We have shown that similarity based on word vectors beats the classical approach with a large margin, whereas other vector representations of senses and sentences fail to even match the classical baseline. Further, we have presented thorough error analysis and reasoning to explain the results with appropriate examples from the dataset for deeper insights.
\end{abstract}

\maketitle

\section{Introduction}
\label{intro}
In the recent years, a considerable amount of work has been done towards ethical aspects of AI \cite{zemel2013learning, feldman2015certifying}. Majority of such efforts focus on identifying and removing bias from the datasets, training process and the trained models \cite{NIPS2017_6988,kamishima2011fairness}. This literature assumes that the information about the sensitive features \cite{zadrozny2004learning} and implications of biased decisions are known in advance.

But in general setting, all the implications of abiding laws are not known up-front and detailed manual study of the large set of legal documents applicable for the domain is needed before attempting to de-bias the machine learning system as per the legal constraints. This situation demands a system or algorithm that can analyze all relevant legal documents to identify sentences or policies that are pertaining specifically to concepts like fairness, bias and discrimination. Typically, legal-domain sentences are long and complex \textit{e.g.,} consider the following sentence taken from  \href{https://www.law.cornell.edu/uscode/text/42/2000e-2}{US code}\footnote{https://www.law.cornell.edu/uscode/text/42/2000e-2}:

\begin{quote}

\textit{``It shall be an unlawful employment practice for an employment agency to fail or refuse to refer for employment, or otherwise to discriminate against, any individual because of his race, color, religion, sex, or national origin, or to classify or refer for employment any individual on the basis of his race, color, religion, sex, or national origin.''}
\end{quote}

These legal documents are written in rigorous fashion in order to achieve robustness and to remove chances of ambiguity. As a side-effect, often, these sentences tend to become complex and hard to interpret even for most humans, especially those who do not have enough background knowledge of legal domain. A recent proposition by \cite{shaikh2017end} highlighted the importance of approaches for analyzing such complex documents for ensuring fairness as part of a high level end-to-end system architecture, but they did not mention any specific method that could address this issue.

Most obvious first step for automatically interpreting or understanding any sentence is to parse the sentence and identify dependencies among the syntactic components. But unfortunately, parsing long and formal sentences is cumbersome as well as time consuming and needs a lot of memory \cite{chen2014fast, klein2003accurate} even for the best parsers. Hence we decided to address the problem with alternative means \textit{viz.,} Wordnet \cite{fellbaum1998wordnet} based semantic relatedness \cite{pedersen2004wordnet} and vector representations of words \cite{mikolov2013efficient} and sentences \cite{le2014distributed,mikolov2013distributed}. The reason for choosing these two categories of techniques is to study the relative effectiveness of classical NLP based techniques and the recent vector representation based techniques.

On the classical side, we populated a set of seed senses from Wordnet that are related to concepts like fairness, bias, discrimination \textit{etc.} We computed Wordnet based similarity of each word in the candidate sentences with the seed senses. If the maximum similarity 
is above threshold, we marked the candidate sentence as a fairness policy. We used this as a classical baseline for evaluating the family of vector based approaches.

Even though many recent experiments have shown that vector based approaches show promising results for various tasks, directly using them for applications relying heavily on semantic relatedness without any added computations may not work in all cases due to various reasons including but not limited to weaker adaptability across domains and the coarse grained nature of relatedness captured by them. We have shown through our experiments that the baseline of Wordnet similarity based classical approach is hard to beat for many kinds of vector representation based approaches which work on word and sentence level of granularity. We have finely analyzed the error cases for each approach to point out the strengths and weaknesses of all the difference approaches that we have tried. 

Once such potentially relevant fairness policies are extracted, it becomes much easier to study their compliance for the target machine learning systems. We have studied this task of fairness-policy extraction from two independent perspectives \textit{viz.,} classical NLP approaches and vector based approach. Further, we performed error analysis on the results which reveals strengths and weaknesses of both the approaches with respect to the task of sentence extraction of a given semantic category.

Rest of the paper is organized as follows, section \ref{related} compiles the various kinds of approaches proposed in the past for solving similar and related problems. Even though we could not find any direct approach for extracting fairness policies, there are efforts targeting related problems like sentence similarity. Section \ref{classical} gives a detailed description including the implementation details for the classical NLP based approach. The next section, (section \ref{embedding}) describes the vector based approaches along with their experimental setup and work flows. Section \ref{Dataset} describes the collection and usage of the dataset that we have used for this task. This is followed by  section \ref{results}, that covers the results obtained for each of the approaches. Section \ref{errorAnalysis} provides a deep insight into the outcomes obtained for different parameter values in order to analyze them better. Section \ref{discussions} provides a comparative analysis of both the approaches, followed by a conclusion and possible future directions covered in section \ref{conclusion}.
\section{Related Work} \label{related}
Strictly speaking about extraction of fairness policies from the legal domain corpora, not much has been explored in the literature towards the exact task that we are attempting. In fact, \cite{roehling1993extracting} argues against such practices due to possible real-life impact it can have due to errors introduced in the automated extractor. Nevertheless, we can find similar efforts of policy extraction in legal domain with some assumptions which can be considered for solving the problem statement that we established in section \ref{intro}, which are covered in the next paragraph.

 \cite{zou2017port} proposed a method to represent the rules in weakly structured English in the structured form for automated decision making. The approach proposed by \cite{NavasLoro2017MiningRA} targets legal domain text but aims at extracting temporal events for reasoning. Even for extracting the events, authors did not propose any custom novel method and instead suggested a combination of existing tools that could perform the annotations on the source text for further reasoning.

With a slight relaxation on the legal domain and extraction of fairness policies, we have many good generic approached which try to classify or rank the sentences for a particular objective based on semantic interpretation of the sentences.

In the medical domain, \cite{agarwal2009automatically} worked towards classifying medical domain sentences into various rhetorical categories like introduction, method, results and discussion. Similarly \cite{mcknight2003categorization} performed sentence classification on medical corpus targeting only two categories \textit{viz.,.} structured and unstructured abstracts. This particular kind of sentence classification looks similar to policy extraction in terms of identifying sentences belonging to a particular semantic category but the key difference lies in the granularity of the semantic categories used. `Fairness policies' is a very narrow and specific semantic category as compared to the categories considered by the above approaches. Thus very targeted relatedness computation is needed to establish belongingness to the class of `Fairness Policies'.

As discussed in the above paragraph, the task of `fairness policy extraction' can also be looked at as a classification problem where the two classes would be `fairness policies' and `non-fairness policies'. It is tough to train a fully supervised classifier due to lack of labeled domain-specific training data for this task. Hence our best bet for now is to go for semi-supervised approaches motivated from bootstrapping \cite{yarowsky1995unsupervised}. The key idea in bootstrapping is to start with a small seed set of labeled examples and tag the large untagged dataset by finding the similarity of each data point with the seed set representing each class. In our case, sentences are the data-points and there are various ways in which we can capture the similarity or relatedness among sentences.

broadly speaking, the methods of computing sentence similarity methods can be categorized into two major categories \textit{viz.,} classical NLP approaches and vector representation based approaches. classical approaches rely mainly on semantic dictionaries like Wordnet \cite{fellbaum1998wordnet} or the distributional similarity \cite{lee1999measures}, whereas more recent vector representation based approaches rely on capturing the contextual features to learn the fixed length representations of words \cite{mikolov2013efficient}, senses \cite{trask2015sense2vec} and even larger compositions like phrases, sentences paragraph or even full documents \cite{mikolov2013distributed,le2014distributed}. The notion of similarity captured by different methods mentioned above greatly varies and must be understood before using them for specific applications. We have tested their effectiveness for solving our problem statement of policy extraction. Our experiments show what aspects of candidate sentences are captured by these techniques and provide a hint towards further improving these techniques for solving similar problems more effectively.



\section{Background} \label{background}

As established in the previous section, semantic relatedness of words, senses and sentences is the key idea we need to rely on for extracting the fairness policies. Let us take a quick dive into different approaches of semantic similarity mentioned in the related work.

\subsection{Classical approaches of Semantic similarity}

This category includes many popular measures which mainly rely on Wordnet gloss, information content, path based measures or their combinations. Some well known approaches from each sub-category are briefly enumerated below.

\subsubsection{Path based similarity} These approaches rely on the discovered relationships among concepts based on either the shortest path or a path following some specific constraints of directed edges and depth from the root of the is-a hierarchy. \cite{leacock1998combining} computes the similarity by finding the shortest path between the two concepts and normalizing it with the longest possible path in the whole hierarchy of the Wordnet. Whereas, \cite{wu1994verbs} computes the similarity by checking how far are the candidate concepts from their lowest common ancestor in the hierarchy. The similarity is computed by finding the depth of the lowest common ancestor and normalizing them with the average depth of each candidate concepts.

\subsubsection{Similarity based on information content:} This category of approaches use both Wordnet hierarchy and a large corpus to figure out the similarity between two concepts. All the approaches of this kind primarily rely on the information content of lowest common ancestor of the candidate concepts. If the lowest common ancestor is highly specific, the similarity among the concept will be higher and vice versa. The pioneering work of this kind was put forth by Resnik \cite{resnik1995using} where he showed that information content based similarity computed using brown corpus outperforms the baseline of simple probability based similarity and path based similarity.

In general, if we talk about relatedness among the concepts (not restricted by the part-of-speech categories) instead of mere similarity, there are many other good approaches worth mentioning like adapted Lesk \cite{banerjee2002adapted}, but with our problem statement into mind, those generic relationships may not be very useful to us.

\subsection{Vector representation based approaches to semantic similarity} 

Starting with Word2Vec \cite{mikolov2013efficient}, there are innumerable successful approaches which can represent components of natural language text in the form of fixed length vectors. We can find plenty of approaches that can represent words \cite{mikolov2013efficient,pennington2014glove}, senses \cite{trask2015sense2vec}, phrases \cite{mikolov2013distributed}, sentences, paragraph and documents \cite{le2014distributed}. One common and useful thing about all these approaches is that the similarity among the vector representation gives a good estimation about the semantic relatedness of the original text components.\\

\subsubsection{Vector representation of words:}~\\

\noindent One of the most noteworthy approach for representation of words is Word2Vec \cite{mikolov2013efficient}. They learn neural network with a single hidden layer that can predict the context given the word (skip-gram model) and word given the context (CBOW) model. The corresponding rows hidden layer of these networks are the vector representations of the words. These representations depict an interesting property that words occurring in the similar contexts have similar vector representations. This is a very useful property that we can leverage for our problem statement. 

Glove (Global Vectors) \cite{pennington2014glove} is another interesting approach of this kind. They demonstrated that their representation outperformed both the formulations of Word2Vec in the word analogy task despite being more efficient in terms of time complexity. Instead of relying on the ability to predict the context words or the missing words as in Word2Vec, Glove generates their representation directly by analyzing the n*n matrix of co-occurrence probabilities.\\

\subsubsection{Vector representation of senses:} ~\\

\noindent One common drawback of the approaches that represent a words with unique vectors is that they cannot distinguish among the multiple senses that a word could take in different contexts. \cite{trask2015sense2vec} addressed this issue by proposing a modified representation that can disambiguate the sense of the word and return the representation for that specific sense. On the face at least, it looks promising for our task.\\

\subsubsection{Vector representation of phrases:} ~\\

\noindent Another drawback of the word-centric representation is that they cannot represent the joint meaning of multi-word expressions. \cite{mikolov2013distributed} addressed this issue by identifying common phrases using co-occurrence based technique and replacing them by a unique token throughout the corpus before training. \\

\subsubsection{Vector representation of sentences, paragraphs and documents: } ~\\

\noindent Le and Mikolov \cite{le2014distributed} extended the framework of Word2Vec \cite{mikolov2013efficient} by introducing `document id/paragraph id/sentence id' as yet another input to the CBOW and skip-gram networks of Word2Vec. The weight matrix of the newly added input trained in the process of learning acts as a set of sentence/paragraph/document vectors for the input text. They evaluated the sentences vectors on sentiment analysis task and showed that it performed better than state-of-the-art methods. Sentiments are very coarse grained categories compared to the category of `fairness policies' in our problem statement in our problem statement, but it will be worthwhile to try out this approach for our task.
\section{Task description and Motivation}
\label{problem}

Before looking into possible solutions and the experiments, it is essential to establish a crisp problem statement that we are trying to address. Shaikh et al., \cite{shaikh2017end} highlighted the need of an end-to-end machine learning platform that ensures fairness. They proposed a high-level architecture that can interpret the relevant legal documents to ensure that underlying machine learning flows are compliant with the fairness policies from the legal documents. But they did not provide any methods or concrete solutions to realize each individual component of their system, which also include interpreting the fairness constraints from legal policy documents. Identifying the exact subset of fairness policies from such documents can greatly speedup the entire process of ensuring fairness in both automatic and manual settings.

With this motivation in mind, we now formally define our problem statement as: 

\begin{quote}
``Given a set of legal documents, automatically identifying all the sentences or policies that are meant to enforce fairness among various protected groups in a particular context''.\end{quote}

Here, every sentences from the input legal documents is considered as a potential fairness policy and evaluated for filtering. For simplifying the solutions, we are assuming that all the sentences are independent of each other. This may not be the case always, but should be fine as far as our problem statement is concerned. Even if the extracted fairness policy is linked to other sentences, after filtering we can always go back to the source document  to complete it's meaning. But we believe that given the formal nature of legal documents, such occurrences would be rare.

The most natural way to address this problem could be to learn a supervised classification for sentences. But as discussed earlier, we do not have any publicly available large dataset in the legal domain that has explicitly tagged fairness policies. Thus we are left with the choice of semi-supervised approach motivated by \cite{yarowsky1995unsupervised} where we start with a small set of manually tagged examples and grow the tagged set progressively by finding similarity of the untagged candidate sentences with the seed set. With this principle in mind, we have tried with various ways of computing similarity among the candidate sentences and the seed words/senses/sentences and compared their effectiveness backed by the thorough error analysis.
\section{Classical Approach} \label{classical}

In the section \ref{background}, we discussed about two categories of classical approaches for computing semantic similarity among two senses. We can use these similarity metrics to determine if a given sentence in indeed a fairness policy. 

\subsection{Creating the set of seed senses}
To represent the class of fairness policies, we manually created a seed set of Wordnet senses which can be used as a reference for similarity. This seed set is created as follows:
\begin{enumerate}
\item Start with the small set of words which when appear in any sentence strongly endorse the belongingness to the class of `fairness policies'. \textit{E.g.,} fair, discriminate, preferential, bias \textit{etc}.
\item Manually identify the correct sense of each word in the set that we created in the previous step.
\item Grow the set of senses from the previous step by finding all the senses that have very high similarity with the original set of senses. The threshold of the similarity is maintained high in order to restrict the number of seed senses below 30. We noticed that after 30, a slight topic drift was happening.
\end{enumerate}

With this set of seed senses defining the sentence type that we want to search, we are good to go ahead and score the candidate sentences.

\subsection{Method details}
In this approach, we classify a sentence as fairness policy only if at least one of the words in the sentence is strongly related to the fairness. In other words, if at least one word from a given sentence has a high similarity score with any of the seed sense, we mark the sentence as a fairness policy.

We compute the similarity of a word $w_i$ with a seed sense $s_j$ as follows:

$$sim(w_i, s_j) = \max_{s_k \in senses(w_i)} sim(s_k, s_j)$$
\\
Both the classical methods of computing semantic similarity (path based and information content based) essentially rely on specificity of the lowest common ancestor of the two concepts in the Wordnet's is-a hierarchy. They differ only in the way they determine the specificity. The path based approaches use path lengths in the Wordnet itself to determine the specificity \textit{e.g.} \cite{wu1994verbs}, whereas information content based approaches rely on probability of occurrence of the lowest common ancestor in a large sense tagged corpus to estimate its specificity. While both the approaches have there strengths and weaknesses, they largely perform in a very similar manner in context of our problem statements. Thus we decided to experiment with only path based similarity \cite{wu1994verbs} among two categories as a representative of classical approaches. We performed POS-tagging on the sentences before computing the similarities in order to reduce the number of candidate senses that a word can take.

\section{Vector Representation Based Approaches} 
\label{embedding}

Recently emerging family of approaches that represent the components of text (like words, senses, phrases, sentences, paragraphs or even documents) in the form of fixed length vectors are gaining rapid popularity due to their versatile applicability to many problems in NLP. Most of these approaches look promising for our problem statement.

\subsection{Similarity of Word Vectors}

On a very broader note, two very well knows approaches for vector representation of words \textit{viz.,} GloVe \cite{pennington2014glove} and Word2Vec \cite{mikolov2013efficient} estimate the representation of words based on the context in which a particular word occurs. There are, of-course, differences in the way they capture the contextual clues. GloVe relies on the co-occurrence probabilities of words whereas Word2Vec relies on the ability to predict the context from the word or vice-versa for coming up with the vector representation.

GloVe claims a slight advantage over Word2Vec in terms of time and space complexity. Given the similarities and differences among these two approaches of same category, we have chosen to try out GloVe for attempting our task.

We have chosen 5 seed words for applying GLoVe \textit{viz.,} discriminate, fairness, discrimination, justice and bias. The vector representations of these words are used as a seed set for testifying the sentences. Since word vectors do not differentiate between multiple senses a word can take, we can only specify words in the seed set and not their relevant senses.

Similar to the classical approach, we have considered the candidate sentences as a bag of words, and marked it as a fairness policy if at least one of the word vector return high similarity with any of the seed vector. 

A pre-trained model named `en\_core\_web\_sm' shipped with SpaCy \cite{spacy2} was used to get the vectors for words. This model is trained on web data including blogs, news, and comments.

\subsection{Similarity of Sense Vectors}

Despite of being successful at various tasks, word vectors lack the ability to represent different senses of the same word. Thus, it makes sense to try out Sense2Vec \cite{trask2015sense2vec} to see if using the sense disambiguated representation makes any positive difference towards the task of fairness policy extraction.

We used the manually chosen correct relevant senses of the same 5 words that we chose for GloVe as the seed senses and computed the similarity in a very similar way to that of GloVe. The key difference here was, that we used word senses as the basis of similarity instead of words themselves.

\subsection{Similarity of Sentence Vectors}

Vector representations like Para2Vec a.k.a. Doc2Vec \cite{le2014distributed} enable us to represent the semantics of larger chunks of texts like sentences, paragraphs and documents as fixed length vectors. These vectors capture the semantics of the full chunk of text that they represent. Thus, similarity between such vectors can be very helpful for our task.

We used following set of five known fairness policies as the seed set:
\begin{enumerate}

\item \textit{It shall be an unlawful employment practice for an employer to discriminate against any of his employees or applicants for employment, for an employment agency to discriminate against any individual, or for a labor organization to discriminate against any member thereof or applicant for membership, because he has opposed, any practice made an unlawful employment practice by this title, or because he has made a charge, testified, assisted, or participated in any manner in an investigation, proceeding, or hearing under this title.}
\item \textit{It shall be an unlawful employment practice for an employer, labor organization, or employment agency to print or publish or cause to be printed or published any notice or advertisement relating to employment by such an employer or membership in or any classification or referral for employment by such a labor organization, or relating to any classification or referral for employment by such an employment agency, indicating any preference, limitation, specification, or discrimination, based on race, color, religion, sex, or national origin, except that such a notice or advertisement may indicate a preference, limitation, specification, or discrimination based on religion, sex, or national origin when religion, sex, or national origin is a bona fide occupational qualification for employment.}
\item \textit{The National Policy on Education (NPE) is a policy formulated by the Government of India to promote education amongst India's people. The policy covers elementary education to colleges in both rural and urban India. This policy calls for "special emphasis on the removal of disparities and to equalise educational opportunity," especially for Indian women, Scheduled Tribes (ST) and the Scheduled Caste (SC) communities.}
\item \textit{The right to religious freedom means that people should not be forced to act against their convictions nor restrained from acting in accordance with their convictions in religious matters in private or in public or in association with others}
\item \textit{It shall be an unlawful employment practice for an employer - to fail or refuse to hire or to discharge any individual, or otherwise to discriminate against any individual with respect to his compensation, terms, conditions, or privileges of employment, because of such individual's race, color, religion, sex, or national origin; or - to limit, segregate, or classify his employees in any way which would deprive or tend to deprive any individual of employment opportunities or otherwise adversely affect his status as an employee, because of such individual's race, color, religion, sex, or national origin."}
\end{enumerate}

With these policies as reference, we trained the para2vec model as follows:

\begin{enumerate}
\item Used 100k news headlines from Kaggle\footnote{https://www.kaggle.com/therohk/million-headlines} as base dataset for unsupervised training of the Para2Vec model.
\item Appended another 165 sentences which we have manually tagged as either `fairness policy' or `not a fairness policy' including five sentences from the seed set. The tags will only be used for evaluation later. The purpose of using them here is to learn their vector representations in the training process.
\item Trained the Para2Vec model in order to get the matured vector representation for 165 sentences of our interest.
\end{enumerate}

We have intentionally chosen three policies having strong hint words like `discriminate', `fair' \textit{etc.} and other two policies not having any of those words for better coverage.

Vector representation of each candidate sentence was then compared with seed vectors to determine if the sentence is similar to the seed fairness policies.

\section{Dataset} \label{Dataset}

The dataset used for all the experiments consists of two parts \textit{viz.,} 165 manually tagged gold sentences with known labels and 100k untagged sentences without any labels . The smaller hand-tagged dataset is used for checking the performance of the various methods whereas the larger untagged dataset is used to assist better learning by the approaches like Para2Vec. Out of 165, 5 fairness policies are used as seed set for Para2Vec, and remaining 160 sentences are used for testing various approaches. 105 out of 160 are fairness policies, and other 55 are n-fairness policies.

The manually tagged gold policies are  taken from multiple legal sources \textit{viz.,} \href{https://consumer.findlaw.com/credit-banking-finance/equal-credit-opportunity-act.html} {Equal Credit Opportunity Act}\footnote{https://consumer.findlaw.com/credit-banking-finance/equal-credit-opportunity-act.html}, \href{http://employment.findlaw.com/employment-discrimination/title-vii-of-the-civil-rights-act-of-1964-equal-employment.html} {Civil Rights Act}\footnote{http://employment.findlaw.com/employment-discrimination/title-vii-of-the-civil-rights-act-of-1964-equal-employment.html},
\href{https://meta.wikimedia.org/wiki/Legal/Legal_Policies}{Fundamental Rights}\footnote{\url{https://en.wikisource.org/wiki/Constitution_of_India/Part_III}}. These sentences are mix of employment and labor laws, civil rights and equal credit opportunity laws. Note that even if we have chosen the laws from the categories strictly related to fairness, not all the sentences are directly related to ensuring or defining fairness.

For all the the vector based approaches except Para2Vec, no additional training data was needed, since good quality pre-trained models were available. But for Para2Vec, it was essential to train the model with 165 policies along with many other sentences from this domain. Thus, we collected 100k sentences from various legal domain sources
For the training purpose in Para2Vec, we trained the model using 100k news headlines from \href{https://www.kaggle.com/therohk/million-headlines} and 160 policies from various different sources mentioned above.

\section{Results} \label{results}
All the approaches were evaluated against 160 manually tagged policies. We ran four experiments for the fairness policy extraction task using classical path based similarity, GloVe, Sense2Vec and Para2Vec. Experiments revealed that GloVe yielded the best performance among all the approaches both in terms of F1 score as well as the area under the ROC curve \cite{david2007roc}.

\smallskip
ROC Curves between the specificity and sensitivity values have been plotted. Each ROC graph consists of two curves, one for each class. Even though areas under the curves of both the classes in the same graph are same, we have shown both the graphs for the sake of completeness.

\begin{table}[b]
\small
\caption{Combined Results \\}
\label{Combined Results}
\begin{center}
\begin{tabular}{|c|c|c|c|c|}
\hline
Approach & Macro P & Macro R & F1 & AUC \\
\hline
\small Classical Approach & 0.651 & 0.656 & 0.653 &  0.64\\ \hline
\textbf{GloVe} & \textbf{0.720} & \textbf{0.710} & \textbf{0.715} & \textbf{0.78}\\ \hline
Sense2Vec & 0.641 & 0.631 & 0.636 & 0.63 \\ \hline
Para2Vec & 0.501 & 0.501 & 0.501 & 0.46\\ \hline
\end{tabular}
\end{center}
\end{table}

\subsection{Classical Approach}
All the metric scores were recorded for a range of threshold values from 0.1 to 0.9 with the step of 0.1. The best Case performance of the approach was obtained for the threshold Value of 0.8 giving F1 score to be 0.653.

\begin{figure}[b]
  \includegraphics[width=\linewidth]{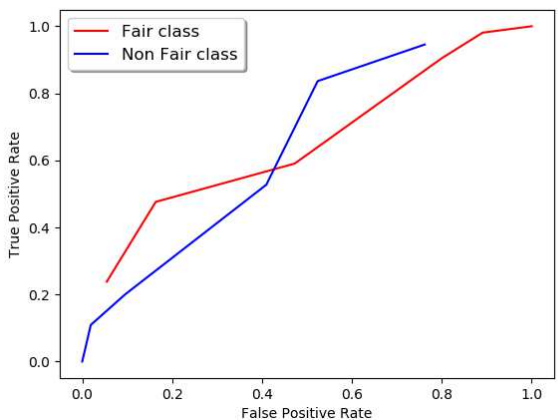}
  \caption{ROC (Classical Approach)}
  \label{classicalROC}
\end{figure}

As evident from Fig \ref{classicalROC}, descent amount of area (0.64) is under the curve indicating that the classifier is acceptable.

\subsection{Vector Based Approach}
The three different kinds of approaches relying on vector representation yielded highly varied results.
\newline

\subsubsection{GloVe}
\begin{figure}[b]
  \includegraphics[width=\linewidth]{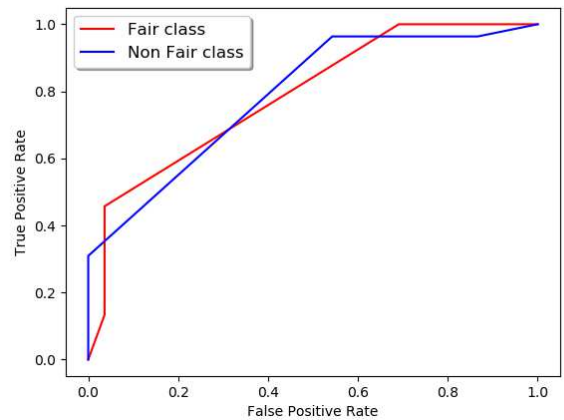}
  \caption{ROC (GloVe)) }
  \label{GloVeROC}
\end{figure}

~\\
Similar thresholds were tried for this approach ranging from 0.1 to 0.9 with the steps of 0.1. This approach returns best F-score at the threshold value of 0.4, with macro recall value being 0.710 and macro precision of 0.720. Table \ref{Combined Results} provides the values of different scores recorded. ROC plot in Fig \ref{GloVeROC} explains the relationship between sensitivity and specificity for both the classes Fair and Non-Fair. As we can see, ROC graph for GloVe has got largest area under the curve (0.78) which is a very good number given the semi-supervised nature of approached that we have tried.
\newline

\begin{figure}[b]
  \includegraphics[width=\linewidth]{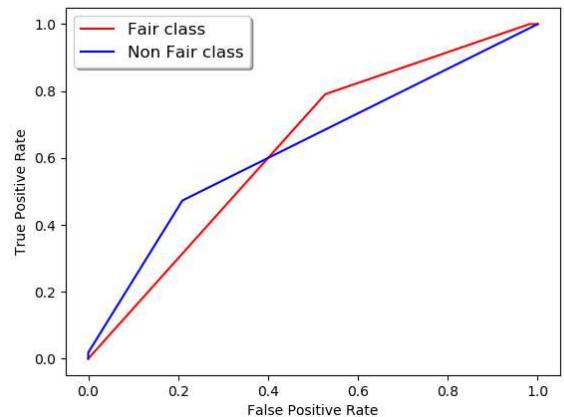}
  \caption{ROC (Sense2Vec) }
  \label{Sense2VecROC}
\end{figure}

\subsubsection{Sense2Vec}~\\
Again, for Sense2Vec as well, we used same set of threshold values  0.1 to 0.9 with the steps of 0.1. The best threshold point is 0.6, which provides the macro recall 0.631 and 0.731 macro precision score. The relationship between the True Positive Rate and False Positive Rate for both the classes is depicted in the Fig \ref{Sense2VecROC} through an ROC curve. To our surprise, despite of modeling the sense specific embeddings, Sense2Vec could not beat GloVe as could barely match with the classical baseline. More error analysis is done in the subsequent sections.
\newline

\subsubsection{Para2Vec}
\begin{figure}[b]
  \includegraphics[width=\linewidth]{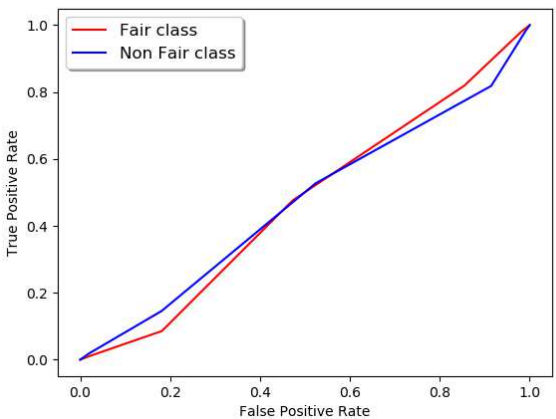}
  \caption{ROC (Para2Vec) }
  \label{Para2VecROC}
\end{figure}

~\\
The best threshold for Para2Vec came out to be 0.5 with F1 score of 0.501. The ROC plot in Fig \ref{Para2VecROC} is largely linear indicating not so good performance. The area under the ROC curve is 0.46. AUC (Area under the ROC curve) below 0.5 is generally not considered good. Even though Para2Vec models overall semantics of the sentences, it performed poorly. As per \cite{le2014distributed} it performed very well with classification of sentences into sentiment polarities. One possible reason for low performance could be the nature of the classes in our task which are not as wide as the sentiment categories and do not go very well with the coarse grained semantics captured by this approach.

\section{Error Analysis} \label{errorAnalysis}
Each approach has been analyzed for the best threshold and some key insights have been laid down for each of the proposed approach, explaining the possible reasons for the erroneous results.
\subsection{Classical Approach}
The following inferences were drawn from the manual analysis of the results belonging to Classical Approach.
\subsubsection{False Negatives}
\begin{itemize}

\item Words such as \textbf{unlawful}, \textbf{equal} even in the right context, pose a low similarity score, thus unable to clear the threshold barrier.


\item Certain sentences contain an implied meaning relating to the fairness even though it may not contain any explicit word for the same. Such sentences are not identified by this approach.
\end{itemize}
\subsubsection{False Positives}
\begin{itemize}

\item
The words discrimination and fair appearing in different context incorrectly causing the sentence to be classified as a fairness policy.

\item Various words like `supervision', `enforcement' are highly related to the word `discrimination' provided in the seed senses dictionary due to which they present a high similarity rate with each other even though they may not relate to the fairness context.
\end{itemize}

\subsection{Vector Based Approach}
The following reasonings were drawn from the manual analysis of the results with the available ground truth: 
\subsubsection{Global Vectors (GloVe)}
\subsubsection*{False Negatives}
\begin{itemize}

\item Sentences which semantically represent fairness related issue, are not reflected as fairness policies since words in those sentences have no similarity with the seed words.


\item Words such as \textbf{legal} and \textbf{equal} are ignored by this algorithm as they have a similarity score of 0.4 with the set of seed words and thus couldn't reflect as fairness policy since its threshold is greater than 0.4.

\end{itemize}
\subsubsection*{False Positives}
\begin{itemize}

\item Words which are similar to any one of the seed words for example, \textbf{civil} is similar to word \textbf{discrimination} categorizes the policy as a fairness.

\item Words like discrimination are right away causing the classifier to tag the sentence as a fairness policy, fails to recognize the context in which it occurs.

\item Few vectors are highly similar to each other as per the model they have been trained upon and eventually create an exception.
For example: `employment' and `discrimination' have a similarity score of 0.54.

\end{itemize}

\subsubsection{Sense2Vec}
\subsubsection*{False Negatives}
\begin{itemize}

\item At certain point, the approach fails to work for the sentences that have implied meaning hidden.
One such example of a policy is \textbf{It aims to curb black money}, this policy talks about fairness on a very high level but doesn't contain any explicit terms to trigger any of the word or sense based approaches.


\end{itemize}
\subsubsection*{False Positives}
\begin{itemize}

\item  Seed words present in the sentence but taking some other sense which is not distinguished by the approach.
Example- `They shall make such further reports on the cause of and means of eliminating discrimination'.

\item Few vectors are highly similar to each other as per the model they have been trained upon and eventually create an exception.

\end{itemize}

\subsubsection{Para2Vec}
\subsubsection*{False Negatives}
\begin{itemize}
\item Para2Vec failed to identify even the sentences containing the phrase `discriminate against ...' which is a very strong indicator for being a fairness policy but hard to capture on the high level, given the length of the sentence.
\end{itemize}
\subsubsection*{False Positives}
\begin{itemize}
\item A sentence containing definition of employment agency was classified as a fairness policy possibly because it picked employment as one of the potential clue from the seed set. This does not mean that Para2Vec is doing its job incorrectly, but the way we are trying to use it may not be the right way. We may have to either modify the way in which we can leverage it in the best way or slightly change the process of training the model that can capture the required clues correctly.
\end{itemize}

\section{Discussions} \label{discussions}
In this section, we summarize all the findings and highlight on the interesting insights from the error analysis that could be useful for more efforts into the same task and even for other similar tasks.

\begin{figure}[h!]
  \includegraphics[width=\linewidth]{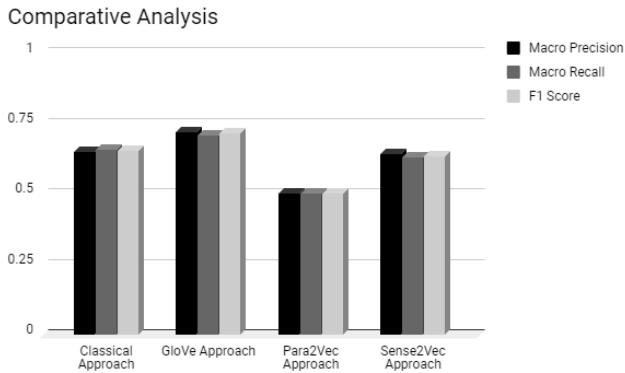}
  \caption{Comparison of various approaches}
  \label{fig:Comparison}
\end{figure}

Figure \ref{fig:Comparison} clearly shows that GloVe, vector based approach outperforms all the other approaches with the large margins. Whereas, to our surprise, Sense2Vec and para2Vec could not show their full potential in this particular setting due to various reasons which we'll discuss here.

The classical approach based on the path based similarity performed decently without even looking into the sentence composition by merely looking at a sentence as a bag of words. One possible reason for that is the ability to specify exact sense IDs from the WordNet as the seed set. Hence, the only sense ambiguity that we had was with the senses of the words in the candidate senses. We tried performing Word Sense Disambiguation on the candidate senses with off-the-shelf approaches like Adaptive Lesk \cite{banerjee2002adapted} but it didn't help much. There are stronger supervised and unsupervised WSD approaches which we could use, but supervised models are highly domain specific and the unsupervised models are not very highly accurate and need mapping of the discovered word categories with the real senses.

Word vectors (GloVe) performed really well despite all the problems discussed for the classical approach along with the inability to add specific senses in the seed dictionary unlike the classical approach. We used pre-trained GloVe vectors which are not domain specific, but still achieved best performance for this task. There are of course cases where this approach failed due to slightly generic nature of relatedness captured by this approach as discussed in the error analysis, but the number of such cases is not very large.

Sense2Vec captures the context specific representation for words, providing the in-built ability of word sense disambiguation, making us expect more. But unfortunately, it did not work very well in our case. One possible reason could be the difference in the domain of the training data and that of the application. Since GloVe does not at all consider senses, it does not try to capture the domain specific sense distribution, which is not the case with Sense2Vec. Change of domain may have negatively affected Sense2Vec because of change in the underlying sense distribution due to change in the domain. Thus, we should try to re-train the models for such approaches for the new domains before use.

In the end, Para2Vec was also considered as one of the strong potential choice due to its ability to capture high level semantics by not looking at the sentence as a collection of independent words. Fundamentally, this approach was designed for capturing broader level semantic differences like sentiment categories. In our problem statement, on the other hand, the category of fairness policies is very specific and narrow, which is not in correlation with the Para2Vec way of representing text. Capturing such semantics may need more complex architectures like RNN-LSTM \cite{hochreiter1997long} on top of vector representation of words.

\section{Conclusion and 
Future 
Work} \label{conclusion}
We defined the problem statement of automatically extracting 
fairness policies from legal documents motivated from the Fair-AI point of view. We also highlighted the complexity involved in the task given the lack of descent sized training data. Thus, we came-up with the semi-supervised strategies to address this problem using various available methods of semantic similarity and relatedness.

GloVe vectors performed very well for this task despite of not being able to disambiguate the senses of word occurrences and not being able to directly model the sentence level semantics. GloVe outperformed the classical baseline of path based similarity with the large margin. On the other hand, Sense2Vec and Para2Vec, despite of being able to model senses and high level semantics respectively, could not really help much in this task due to various reasons including cross-domain usage of trained models and difference in the granularity of semantics captured by them. The detailed error analysis is presented with the failed examples to support the reasoning.

We could not perform Word Sense Disambiguation (WSD on the candidate sentences due to reasons like unavailability of legal domain sense tagged corpora required for supervised WSD. Using the fine-tuned WSD approach as part of the pipeline is the most obvious next thing to be tried out as future work.

Even though we tried to capture the sentence level semantics using Para2Vec which did not work well, we should try other ways to capture the required kind of semantics by slightly modifying Para2Vec or other ways. As discussed in the previous section, RNN-LSTM on top of the word vectors could be a good choice to start with. We refrained from parsing the sentences due to their length and complexity, but it would worthwhile to attempt semantic understanding via shallow parsing or using some heuristics to parse long sentences with some approximations suitable for our task.

\bibliographystyle{ACM-Reference-Format}
\bibliography{acmart}

\end{document}